# Requirements for Game-Based Learning Design Framework for Information System Integration in the Context of Post-Merger Integration

Ksenija Lace[1][0009-0004-0435-6829] and Marite Kirikova[1][000-0002-1678-9523]

[1] Riga Technical University, Faculty of Computer Science and Information technology, 6A Kipsalas Street, Riga, LV-1048, Latvia
`ksenija.lace@rtu.lv, marite.kirikova@rtu.lv`

**Abstract.** Post-merger integration states unique challenges for professionals responsible for information system integration aimed on alignment and combination diverse system architectures of merging organizations. Although the theoretical and practical guidance exists for post-merger integration on the business level, there is a significant gap in training for information system integration in this context. In prior research specific methods AMILI (Support method for informed decision identification) and AMILP (Support method for informed decision-making) were introduced for the support of information system integration decisions in the post-merger integration. But during the practical application was reported high learning curve and low learner motivation. This paper explores how game-based learning design can address these limitations by transforming static method training into engaging learning experience. The study analyzes foundational learning theories, cognitive load and motivation models, and serious game design frameworks to identify the essential requirements for a game-based learning design framework tailored to information system integration in post-merger integration. Requirements are structured in two components: the transformation process and resulting learning experience. The paper concludes with a plan for developing and evaluating the proposed framework through iterative design and real-world validation.

## 1 Introduction

Mergers and Acquisitions (M&A) are among the most frequently chosen strategies for organizational growth. If executed successfully, they enable the merging parties to create synergies and achieve outcomes that neither organization could accomplish individually [1]. But establishing the synergy requires the implementation of a newly created combined organization, integrating structures, functions, and resources of M&A participants. This new organization should be carefully designed, so that different parts can supplement and strengthen each other when combined, but all duplicates and redundant parts are decommissioned. The process of physical reconstruction of merging organizations is called post-merger integration and is mentioned as one of the key enablers for M&A initiative outcomes [2].

As information systems nowadays play a crucial role in supporting all processes in the organization, it is important to completely identify information systems to be combined, as well as select the best type and extent of combination. Integration of information systems should support decisions made for business architecture, seeking synergies and removing redundancies [3].

With M&A being widely used for a noticeable time, there is a comprehensive body of knowledge and best practices on how the process should be planned and executed. But it should be mentioned, that existing theory is mostly focused on the business perspective of post-merger integration, leaving less attention to the technological level and specifically to the process of merging two or even more information system architectures [4]. In practice, the task of information system integration is often assigned to IT professionals with no or very limited experience in post-merger integration, with an assumption that the information system integration task in this context is similar to the one usually executed when several information systems should be integrated to support flawless execution of the business process which these systems support at different stages or in different phases [5]. However, the task of integrating information systems in the context of post-merger integration is fundamentally different. First, in the standard system integration process, systems to be integrated are already given, but in the context of post-merger integration, systems are to be identified [6]. Secondly, in standard system integration, integration always means the process of establishing a way for two or more systems to exchange data between them, but in the context of post-merger integration, such integration is only one of the options, where other options to consider are to leave systems as is without any kind of integration, to replace one system with another with or without replacing system adjustments, and even to replace all systems with a completely new system capable of supporting the newly created organization [7]. With limited competence of the involved responsible professionals, information system integration in the context of post-merger integration is often executed as the replacement of all systems in the acquired organization by systems of the acquiring organization, making decisions on the fly when information systems are identified while merging or replacing business units. The process is unstructured and does not follow a specific methodology [8].

In order to address this issue, the authors of this article in the previous research have proposed a support method for information system integration in the scope of the post-merger integration, focusing on two of the three process phases, covering decision identification and decision making, but leaving the execution of the made decision out of scope [9]. Two methods were created to support the identification of groups of information systems to be merged (AMILI), and for each of the identified groups, evaluate possible integration options (AMILP). Both methods are described through process and data perspectives, as well as for each of them a proof of concept for the supporting tool was created to store and process information gathered throughout the process. The methods, with a help of supporting tools, were validated with the help of IT professionals without prior experience in information system integration in the scope of post-merger integration, and their results were compared with those of experienced professionals asked to work on the same case study. Results showed that professionals without the previous experience, with a help of the method and the tool, can achieve the same results as experienced professionals. But as one of the potential improvements mentioned by participants in the post experiment survey was the ease of learning the method and tool usage – provided instructions were hard to follow and understand, as well as detailed long descriptions required time and effort to comprehend. This comment becomes even more valid in the context of real post-merger integration, as usually integration activities have a very limited timeframe allowed and are performed under high pressure and stress level on one side, and with insufficient incentive and motivation on the other side [6].

The authors propose the hypothesis that the challenge of learning complicated serious material with a lack of motivation can be compensated by transforming the learning experience into an interactive game-based learning. In this article, authors research existing approaches that could be utilized to transform the created methods training into serious games. Based on the research findings regarding each existing approach applicability, authors propose the requirements for a game-based learning design framework for information system integration in the context of post-merger integration. These requirements, in the future research, can be used for the design of the framework.

The structure of the paper is the following – in the Methodology section, the scope and content of the research are defined, in the Literature Review section, existing research on educational frameworks, challenges, and gamified learning is explored, in the Requirements for the Framework section, the initial requirements for the game-based learning design framework are stated. In the Conclusions and Future Research section, the summary of the current research results is provided, and the next phase of the research is proposed.

## 2 Methodology

Research described in this paper follows the design science methodology [10], and covers the first two phases of the process – problem identification and definition of requirements for a solution.

First, the problem will be stated and justified by the existing research. The authors start with the problem identified in their previous research – the challenge of learning complicated serious material with a lack of motivation - and validate that this problem is current and not solved using the existing research studies. For this, authors plan to perform a literature review of studies published on the topics "existing foundational learning theories", "difficulty of learning serious material" and "lack of learner motivation" to frame the understanding of the main existing learning theories and challenges in learning complex material. Additionally, authors plan to research the articles published on topics such as "educational methods to minimize cognitive load and increase engagement" and "design of game-based learning and serious games" to identify existing solutions and verify if they can successfully solve all challenges stated by existing learning theories. Based on the performed literature review, the initially stated problem could be detailed or adjusted.

Second, the requirements for the solution are defined. As a solution, authors perceive the framework defining the process of transforming a learning experience into a game-based learning experience. This means that the solution can be seen through two perspectives – the transformation process and the final transformed learning experience – and requirements should be defined for each of these perspectives. To elicit requirements for the transformed learning experience, authors plan to use the existing research on the general learning theories, as well as on blockers and enablers of learning complex material and learner motivation and engagement. For requirements related to the transformation process, authors plan to review existing research in the design of learning experiences as well as existing approaches for game-based learning design. Additionally, authors plan to use existing research on information system integration in the context of post-merger integration to identify specific contextual requirements for both solution parts.

## 3    Literature Review

### 3.1    Existing Foundational Learning Theories

The literature review is performed from three complementary perspectives. First, it explores foundational learning theories and best practices to define characteristics of effective learning experiences. Second, the review investigates two specific issues reported in the initial training evaluation – learning difficulty and lack of learner motivation – to identify the root cause leading to them and how they can be addressed. Lastly, serious games are explored as the potential baseline for the development of the game-based learning design framework.

In order to ground the design of the learning experience in a theory that depicts how people learn, the authors selected the following foundational theories:

- Constructivist Learning Theory [11] – proposes that learners actively construct their understanding through active engagement and not through passive perception of information.
- Experiential Learning Theory [12] – states that learning is the most effective when it follows the cyclical process of experience and reflection.
- Situated Learning Theory [13] – emphasizes that effective learning happens in the real-world contexts where knowledge can be practically applied.
- Transformative Learning Theory [14] – highlights the importance of reflection and new insight integration in the existing mental models.

All these theories collectively propose the following characteristics for an effective learning experience:

- Proactive – learning should be driven by active learner involvement, highlighting the need for the ownership, decision-making and exploration activities.
- Applied – learning should be practical and goal oriented, requiring problem-solving, experimenting and practicing tasks.
- Contextual – learning should be mapped to the real-world scenarios, requiring the clear link for learners between what they learned and where they apply it.
- Reflective – learning should incorporate processing and evaluation of the results, supported by periodic self-assessment checkpoints.
- Progressive – learning should evolve and build upon itself, meaning gradually increasing complexity.

As one of the main aspects related to the increased difficulty of learning new material is named the limited human working memory, which is studied under the Cognitive Load Theory [15]. This theory further identifies three different types of cognitive load applied to the working memory while learning and defines how each of these types affects the learning experience and outcomes.

- Intrinsic load – natural load triggered by the complexity of the material itself. Usually, it is impossible to reduce it, since that would require reducing the extent and depth of the topic we want to learn.
- Extraneous load – additional not required and not useful load caused by poor instructional design and learning experience design itself.
- Germane load – useful load required for the practical application, interpretation and creation of the new knowledge constructs.

Recent research in Cognitive Load Theory introduced the concept of element interactivity, which refers to the degree to which individual elements of learning activity interact and must be processed by the learner simultaneously [16]. In a complex domain as post-merger information system integration with multiple inter-dependent procedural, organizational and technical factors, high element interactivity leads to the significant intrinsic load. Designing learning in such a context requires strategies such as segmentation and pre-training, accompanied by worked examples to reduce unnecessary cognitive effort in early learning stages [15]. This is especially important digital learning which has a higher risk to create more unnecessary extraneous load through user interface, narrative, and interaction complexity [17].

While the Cognitive Load theory provides valuable ideas on how to structure learning to optimize cognitive process, they lack a broader systematic process, that could help professionals transform static learning material into a learning experience. In the existing literature can be found several most-cited process frameworks defining how to design learning experiences in a systematic way:

- ADDIE (Analysis-Design-Development-Implementation-Evaluation) Model [18] – linear framework for the sequential process from needs analysis to post-implementation evaluation.

- SAM (Successive Approximation Model) [19] – agile and iterative framework proposing rapid prototyping based on stakeholder feedback.
- Ten Steps for Complex Learning Model [20] – framework is focused on whole-task learning for complex skill development.
- Backward Design [21] – goal-focused framework starting with identification of desired learning outcomes and only then designing corresponding instructional components.

These frameworks provide structured processes for transforming content into learning experience. But they address mainly the cognitive and instructional design dimensions and do not sufficiently cover the emotional and behavioral aspects of learner engagement. To analyze potential improvements from the perspective of the learner motivation, the authors utilize BJ Fogg Behavior Model [22], stating that high motivation can compensate the high difficulty of the task.

There are several existing theories focusing on the motivational aspect, which could be applied to the motivation in learning environments.

Expectancy-Value Theory [23] states that learners evaluate the value of a task and their chances of success and compare it to the expected difficulty of the task to decide if they want to contribute their effort. If the task is too complicated for the value gained and accompanied by high chances of no success, the engagement and commitment levels of learners will be lower.

Self-Determination Theory [24] proposes the three required components for the intrinsic motivation of the learner:

- Autonomy – sense of control and ownership over the experience. Lack of interactivity and personalization negatively impacts interest levels.
- Competence – feeling of being capable to successfully complete the task and achieve the goal. Learning designed with inadequately big non progressive new knowledge areas exposed to learners and complex concepts introduced without proper preparation reduces commitment.
- Relatedness - connection to the context of activity and relationship with other people (social context). As post-merger context itself adds the challenge of social disorientation and lack of confidence in the new organization context, training could benefit of collaborative activities to increase the motivation for the cooperation between different professionals involved in the post-merger integration activity.

Self-Determination Theory defines motivation as a continuum from amotivation, through extrinsic regulation to intrinsic motivation [24]. Game-based learning supports intrinsic motivation by designing game mechanics that address three psychological needs. Research shows that game features like clear goals, immediate feedback, and voluntary engagement increases learners' perception of autonomy and competence, shifting motivation from extrinsic to intrinsic [25]. This makes game-based learning and serious games is the promising foundation for the designing learning experience [26], [27], as they integrate cognitive structuring with emotional engagement. Serios games (games designed with a primary purpose other than pure entertainment) have been increasingly used in education and corporate training [28]. But design of such games requires a structured approach which can effectively combine instructional design, cognitive science, and game mechanics. In the academic literature can be found several most cited serious game design frameworks, each of them focusing on different aspects of learning, engagement and game structure.

- Mechanics, Dynamics, Aesthetics (MDA) [29] – one of the foundational frameworks in the game design. It decomposes the game experience into three interconnected layers: mechanics (the formal structures and rules of the game), dynamics (run-time behavior that emerges when players interact with game mechanics), and aesthetics (emotional responses in players evoked by dynamics).
- Design, Play, Experience (DPE) [30] – is grounded in the MDA, and design can be perceived as mechanics, play ad dynamics and experience as aesthetics. But this framework for each of layers provides the design elements to consider – learning, storytelling, gameplay and user technology. All together is grounded in the baseline technology element.
- Design, Development, Evaluation (DDE) [31] – references both MDA and DPE and proposes the iterative design process where each iteration sequentially goes through design, development and evaluation.
- Learning Mechanics – Game Mechanics (LM-GM) [32] – not a comprehensive design process framework, but rather a model for mapping learning mechanics with game mechanics.

These frameworks can provide a starting point for development of a game-based learning design framework, which can be enriched with methods for decreasing cognitive load and increasing player motivation.

## 4    Requirements for the Framework

To structure the framework effectively for practical application, the requirements are divided into two categories: those related to the transformation process – which defines how the learning experience is designed – and those related to the final transformed learning experience – which defines how the training is delivered and experienced by learners (see Fig. 1).

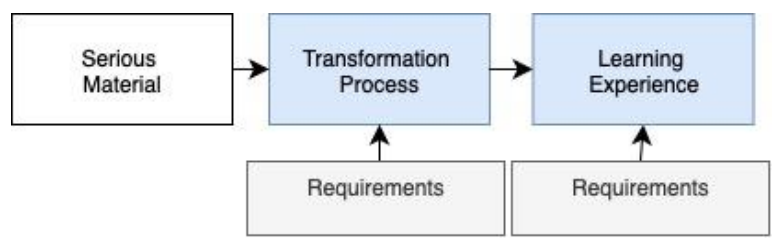

**Fig. 1.** Structure of Framework Requirements

This distinction ensures clarity between the mechanics of creating training and the characteristics of the training itself. To support systematic analysis, all requirements are classified using a common structure [33]:

- Functional requirements describe essential features and capabilities that the framework must support. These at this stage are expressed as high-level requirements and not as specific solution implementations, which will be refined during later stages of the research.
- Quality requirements describe additional characteristics that define how well the framework or experience should perform.
- Constraints identify contextual limitations that must be considered in design decisions.

During the next stages of research multiple learning experiences will be developed based on specific learning goals. As a result, the current requirements for the learning experience are defined in a generic and foundational form, but they will be later tailored in alignment with each specific use case as the framework is applied in practice.

To define the requirements for the transformation process, the following sources were considered based on the performed literature review (see Fig. 2):

- For functional requirements: existing research on instructional design methodologies and established game-based learning design approaches.
- For all types of requirements: existing research on information system integration in the context of post-merger integration.

To define the requirements for the final transformed learning experience, the following sources were used:

- For functional requirements: foundational learning theories, research on difficulties of learning complex material and studies on learner motivation and engagement.
- For all types of requirements: existing research on information system integration in the context of post-merger integration.

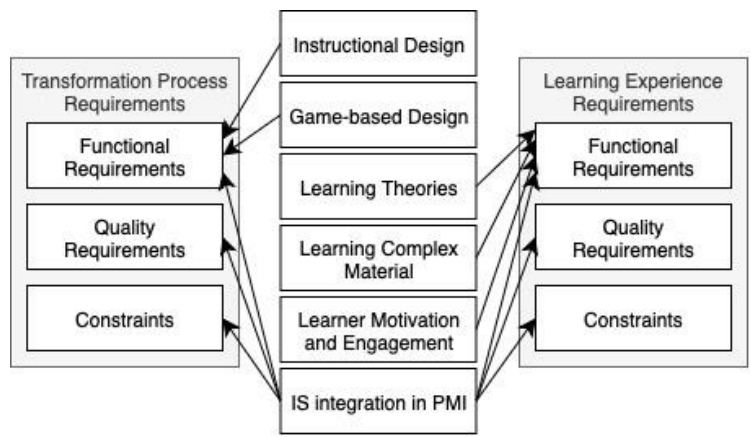

**Fig. 2.** Sources for Framework Requirements

### 4.1    Transformation Process

**Functional Requirements**

1. **ADDIE** - The framework should support a structured, sequential process that guides designers from analysis to evaluation phase
2. **SAM** - The framework should allow for iterative prototyping and continuous feedback loops with stakeholders
3. **Ten Steps** - The framework should enable whole-task learning strategies to build complex skills
4. **Backward design** - The framework should require definition of learning outcomes prior to instructional content development
5. **MDA** - The framework should require definition of game mechanics, prediction of learning dynamics, and intentional design for aesthetics
6. **DPE** - The framework should support design across four layers: learning goals, narrative, gameplay mechanics, and enabling technology
7. **DDE** – The framework should support iterative refinement based on evaluation of learning effectiveness and learner engagement
8. **LM-GM** - The framework should ensure that learning mechanics are effectively mapped to corresponding game mechanics
9. **AMILI/AMILP theory and practice** - The framework should support accurate transformation of AMILI and AMILP methods descriptions into interactive modules for learner training
10. **PMI Stakeholder management** – The framework should allow adaptation of training on role-specific responsibilities and knowledge levels of future learners
11. **Specific PMI challenge management** – The framework should support secure transformation of real-world cases, managing confidentiality and adjusting complexity

**Quality Requirements**

12. **Relevance and adaptability** - The framework should enable tailoring of training experience to different merger types, industries, and legacy systems
13. **Scalability and reusability** - The framework should structure content into modular units to allow replication and extension for multiple merger cases
14. **Reliability and stability** - The framework should consistently support the creation of training that guarantee comparable learning results for different designers
15. **Performance** – The framework should support the rapid design process without delays and breakdowns
16. **Usability and learnability** – The framework should be intuitive and easily learnable by instructional designers
17. **Accessibility** – The framework should comply with inclusive design standards

**Constraints**

18. **Target audience** - The framework should be usable by designers creating training for professionals with no prior experience in educational design and game design
19. **Available time** – The framework should allow training design to be planned and executed under constrained timelines
20. **Technical constraints** – The framework should function within common technical infrastructures and be compatible with existing learning management systems
21. **Organizational constraints** – The framework should align with corporate structures and decision hierarchies
22. **Financial constraints** – The framework should support cost-effective training design using minimal or low-cost resources
23. **Legal and ethical constraints** – The framework should ensure the ethical use of data and compliance with organizational privacy, copyright and confidentiality
24. **Pedagogical constraints** – The framework should ensure learning of core information system integration in the context of post-merger integration concepts and training goals
25. **Content constraints** – The framework should enable transformation of all relevant AMILI/AMILP materials ensuring completeness

## 4.2     Learning Experience

**Functional requirements**

1. **Constructivist Learning Theory** - The learning experience should actively engage learners in constructing understanding through interaction and exploration
2. **Experiential Learning Theory** - The learning experience should cycle learners through concrete experiences, reflection, and conceptualization
3. **Situated Learning Theory** - The learning experience should embed content in realistic PMI integration scenarios to improve relevance
4. **Transformative Learning Theory** - The learning experience should encourage learners to critically reflect on prior assumptions and adapt mental models
5. **Cognitive load theory, Intrinsic load** - The learning experience should match task complexity to the learner's cognitive readiness
6. **Cognitive load theory, Extraneous load** - The learning experience should avoid unnecessary cognitive load through clear design, intuitive UI, and minimal distractions
7. **Cognitive load theory, Germane load** - The learning experience should reinforce practical knowledge construction through varied practice, feedback, and reflection
8. **Self-determination theory, Autonomy** - The learning experience should allow learners meaningful control over decisions and paths
9. **Self-determination theory, Competence** - The learning experience should scaffold difficulty to build confidence and mastery
10. **Self-determination theory, Relatedness** - The experience should integrate social elements to foster collaborative learning and peer motivation

11. **Expectancy-Value** - The learning experience should clearly communicate the importance and practical value of training activities
12. **Expectancy-Value** – The learning experience should provide tasks that are perceived as achievable with visible reward and progression structures
13. **AMILI/AMILP theory and practice** - The learning experience should accurately simulate the two-step AMILI/AMILP process through applied challenges
14. **Stakeholder management** - The learning experience should include role-based tasks that simulate cross-functional collaboration and stakeholder management activities
15. **Specific challenge management** - The learning experience shall prepare learners to navigate time pressure, ambiguity, data gaps, and conflicting priorities in real PMI contexts

**Quality requirements**

16. **Relevance and adaptability** – The learning experience should support adaptation to diverse industry, organizational and IS contexts
17. **Scalability and reusability** – The learning experience should support a range of group sizes and allow for reuse across different training cycles
18. **Reliability and stability** – The learning experience should ensure consistent delivery and learner performance outcomes
19. **Performance** – The learning experience should function smoothly without delay and support session completion within available time
20. **Usability and learnability** – The learning experience should have clear guidance, user friendly interfaces and minimal onboarding time
21. **Accessibility** – The learning experience should support diverse learner needs, including language, technical literacy, and other special characteristics

**Constraints**

22. **Target audience** - The learning experience should adopt to IT professionals with varying IS integration and PMI knowledge and experience levels
23. **Available time** – The learning experience should adopt to different timeframes available for learning
24. **Technical constraints** – The learning experience should be deployable on common enterprise systems without specialized hardware or software
25. **Organizational constraints** – The learning experience should align with existing training formats and protocols used in the organization
26. **Financial constraints** – The learning experience should be acquirable and maintainable within limited training budgets
27. **Legal and ethical constraints** – The learning experience should ensure confidentiality, data security and compliance with organizational and legal norms
28. **Pedagogical constraints** – The learning experience should achieve learning objectives without oversimplifying or gamifying serious content

29. **Content constraints** – The learning experience should cover all necessary topics, tasks, and materials aligned with AMILI and AMILP methods.

While existing game-based learning frameworks like MDA and others provide foundational models for aligning game design with learning objectives, proposed framework distinguishes in three ways:

- It separates the design process into two components: the transformation process (instructional design) and transformed learning experience (game-based learning experience). This separation enables clear guidance for designers, while also ensuring the final result meets stated goals.
- It is aimed to teach the unique characteristics of post-merger information system integration. Existing general purpose frameworks do not address the specialized decision identification and decision making required in this context.
- It takes into consideration constraints specific to post-merger environments, such as lack of data and data confidentiality, as well as limited timeframes and complex stakeholder management.

## 5      Conclusions and Future Research

This paper identifies and structures the requirements for a game-based learning design framework aimed at supporting information system integration in the context of post-merger integration. This research continues the previous work focused on the development of AMILI and AMILP methods. In this stage of research, the goal is to create more effective and engaging training for IT professionals, who are responsible for complex decisions under time and resource constraints.

The research makes three contributions. First, it formalizes the requirements for the learning experience and transformation process using established instructional design models, game-based learning theories, and domain-specific constraints of the post-merger integration context. Second, it integrates psychological models about cognitive load and self-determination to increase the efficiency of learning and learner engagement. Third, it removes the gaps between generic educational frameworks and the practical challenges of information system integration in mergers and acquisitions.

The defined requirements create a foundation for future work. The next phase of the research will focus on the development of the framework itself. This will include the design of transformation guidelines, and example learning experiences based on these guidelines. These artefacts will tested in iterative cycles of implementation and evaluation using real-world case studies. The effectiveness of the framework will be assessed in terms of learner performance, engagement, and long-term retention of knowledge. Future research will also explore how the framework can be adapted to different merger scenarios, organizational cultures, and technical environments. In the long term the framework could support not only training for information system integration in the context of post-merger integration, but also broader digital transformation and complex system change initiatives across organizations.

1.1.1.9 Research application No 1.1.1.9/LZP/1/24/067 of the Activity "Post-doctoral Research" "Development of a Gamified Tool to Enhance IS Integration Decision-Making in M&A: A Methodology-Driven Training Approach"

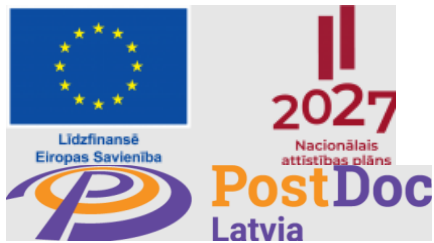